\documentclass[letterpaper, 10 pt, conference]{ieeeconf}
\IEEEoverridecommandlockouts
\usepackage{lineno}
\usepackage{xcolor}
\usepackage{pgfplots}
\pgfplotsset{compat=1.18} 
\usepackage{tikz}
\usetikzlibrary{positioning}
\usetikzlibrary{shapes.geometric}
\usetikzlibrary{shapes.misc}
\usetikzlibrary{external}
\usepackage{paralist}
\usepackage{mathbbol}
\usepackage{amssymb}    
\usepackage{bm}
\usepackage{xcolor}
\usepackage{siunitx}
\usepackage{multirow}
\usepackage{todonotes}
\usepackage{epstopdf}
\usepackage{epsfig}
\usepackage{float}
\usepackage{amsmath} 
\usepackage{amssymb}  
\usepackage{mathtools}
\usepackage[acronym,nomain,nonumberlist]{glossaries}
\usepackage{url}
\usepackage[ruled,vlined,linesnumbered,noresetcount]{algorithm2e}
 \usepackage{algpseudocode}

 \newlength\fwidth

 \usepackage{pgfplots}
\pgfplotsset{compat=newest}
\usepackage{tikz}
\usetikzlibrary{positioning}
\usetikzlibrary{shapes.geometric}
\usetikzlibrary{shapes.misc}
\usetikzlibrary{external}
\usetikzlibrary{shapes,arrows,patterns}
\usetikzlibrary{arrows.meta}
\usetikzlibrary{plotmarks}
\usetikzlibrary{calc,decorations.markings,math}

\usepackage{placeins} 
\usepackage{tabularx}

\graphicspath{ {pictures/} }
\newacronym{ap}{AP}{Average Precision}
\newacronym{av}{AV}{Autonomous Vehicles}
\newacronym{av2}{AV2}{Argoverse2}
\newacronym{aws}{AWS}{Amazon Web Services}
\newacronym{bev}{BEV}{bird's eye view}
\newacronym{cbgs}{CBGS}{Class-balanced grouping and sampling}
\newacronym{ccl}{CCL}{Connected Component Labelling}
\newacronym{cnn}{CNN}{Convolutional Neural Network}
\newacronym{rnn}{RNN}{Recurrent Neural Network}
\newacronym{gru}{GRU}{Gated Recurrent Unit}
\newacronym{cds}{CDS}{Composite Detection Score}
\newacronym{dnn}{DNN}{Deep Neural Network}
\newacronym{epe}{EPE}{End Point Error}
\newacronym{emc}{EMF}{Ego-motion Flow}
\newacronym{ekf}{EKF}{Extended Kalman Filter}
\newacronym{fd}{FD}{Foreground Dynamic}
\newacronym{fs}{FS}{Foreground Static}
\newacronym{bs}{BS}{Background Static}
\newacronym{fsd}{FSD}{Fully Sparse Detector}
\newacronym{fsf}{FSF}{Fully Sparse Fusion}
\newacronym{gps}{GPS}{Global Positioning System}
\newacronym{hd}{HD}{High Definition}
\newacronym{htc}{HTC}{Hybrid Task Cascade}
\newacronym{kf}{KF}{Kalman Filter}
\newacronym{lidar}{LiDAR}{Light Detection and Ranging}
\newacronym{map}{mAP}{mean Average Precision}
\newacronym{mav}{MAV}{Micro Aerial Vehicle}
\newacronym{mhe}{MHE}{Moving Horizon Estimation}
\newacronym{mlp}{MLP}{Multi Layer Perceptrons}
\newacronym{mvp}{MVP}{Multimodal Virtual Points}
\newacronym{nds}{NDS}{NuScenes Detection Score}
\newacronym{nmhe}{NMHE}{Nonlinear Moving Horizon Estimation}
\newacronym{nms}{NMS}{Non-Maximum Suppression}
\newacronym{pdf}{PDF}{Probability Density Function}
\newacronym{radar}{RADAR}{Radio Detection and Ranging}
\newacronym{rbgs}{RBGS}{Range-balanced grouping and sampling}
\newacronym{rpn}{RPN}{Region Proposal Networks}
\newacronym{sir}{SIR}{Sparse Instance Recognition}
\newacronym{ssc}{SSC}{Submanifold Sparse Convolutions}
\newacronym{ssf}{SSF}{Sparse Scene Flow}
\newacronym{slam}{SLAM}{Simulataneous Localization and Mapping}
\newacronym{soa}{SoA}{State of the Art}
\newacronym{sota}{SOTA}{state-of-the-art}
\newacronym{sph}{SPH}{sparse prediction head}
\newacronym{vvm}{VVM}{Virtual Voxel Mixer}
\newacronym{vfe}{VFE}{Voxel Feature Encoding}

\usepackage{graphics} 
\usepackage{amsmath} 
\makeatletter
\let\NAT@parse\undefined
\makeatother
\usepackage[hidelinks]{hyperref} 
\usepackage[nameinlink,capitalise]{cleveref} 
\usepackage{url}
\usepackage{subcaption} 
\usepackage{booktabs}

\definecolor{TableBlue}{rgb}{0.17,0.49,0.75}
\definecolor{TableWhite}{rgb}{1,1,1}

\title{\LARGE \bf
SSF: Sparse Long-Range Scene Flow for Autonomous Driving
}

\author{Ajinkya Khoche$^{1,2}$, Qingwen Zhang$^{1}$, Laura Pereira Sánchez$^{3}$, Aron Asefaw$^{1}$
\\ 
Sina Sharif Mansouri$^{2}$ and Patric Jensfelt$^{1}$
\thanks{$^{1}$KTH Royal Institute of Technology, Stockholm 10044, Sweden. Corresponding author's e-mail: {\tt\small khoche@kth.se}}%
\thanks{$^{2}$Autonomous Transport Solutions Lab, Scania Group, Södertälje, SE-15139, Sweden}
\thanks{$^{3}$Stanford University, CA-94305, USA. }%
}

\IEEEaftertitletext{\vspace{-1\baselineskip}}
\begin{document}


\maketitle
\thispagestyle{empty}
\pagestyle{empty}

\begin{abstract}
Scene flow enables an understanding of the motion characteristics of the environment in the 3D world.
It gains particular significance in the long-range, where object-based perception methods might fail due to sparse observations far away. 
Although significant advancements have been made in scene flow pipelines to handle large-scale point clouds, a gap remains in scalability with respect to long-range. 
We attribute this limitation to the common design choice of using dense feature grids, which scale quadratically with range. 
In this paper, we propose Sparse Scene Flow (SSF), a general pipeline for long-range scene flow, adopting a sparse convolution based backbone for feature extraction. 
This approach introduces a new challenge: a mismatch in size and ordering of sparse feature maps between time-sequential point scans. To address this, we propose a sparse feature fusion scheme, that augments the feature maps with virtual voxels at missing locations. 
Additionally, we propose a range-wise metric that implicitly gives greater importance to faraway points. 
Our method, SSF, achieves state-of-the-art results on the Argoverse2 dataset, demonstrating strong performance in long-range scene flow estimation. Our code will be released at \url{https://github.com/KTH-RPL/SSF.git}.   
\end{abstract}

\glsresetall

\section{Introduction} 
\label{sec:intro}

\glsunset{lidar}
\glsunset{radar}

3D scene flow is an important component of a perception system in autonomous driving. Given a point cloud scan at a given time $t$, scene flow predicts a point-wise displacement vector towards the next time $t+1$. By providing short-term motion information about the scene, it can serve as a prior for various downstream tasks, including 3d object detection, tracking and moving object segmentation. 

However, efficiently estimating scene flow in real-world autonomous driving scenarios remains challenging, particularly when dealing with a large number of input points. 
Existing approaches \cite{jund2021scalable,zhang2024deflow,vedder2023zeroflow} encode point cloud into \gls{bev} pseudo images and use standard 2D convolution layers to extract features. 
Despite the naturally sparse distribution of \gls{bev} feature grids, these methods treat the grid as dense, applying convolutional filters to all grid cells—both occupied and empty.
This dense processing results in sub-optimal computational performance and high memory consumption, limiting the scalability of these methods to long-range scene flow estimation.

Moreover, existing scene flow benchmarks~\cite{jund2021scalable,wilson2argoverse} in autonomous driving limit their evaluation to within 50 meters range of the vehicle.
One reason for this could be that \gls{lidar} data gets sparse with range. 
Nevertheless, the scene flow ground truth can in principle still be obtained, by leveraging sequential box annotations along a track, computing the difference between box motions and applying it to all the points within the boxes. 
Another possible reason is that motion close to the ego vehicle is most important to consider in existing benchmarks aimed mainly for urban driving scenarios. 
That said, increasing the range of perception algorithms would serve to improve the overall safety of an autonomous system. 
Also, as scene flow provides important priors to downstream modules, enhancing the range of existing scene flow methods is becoming increasingly critical. 

\begin{figure}[t]
\centering
\includegraphics[width=0.9\linewidth]{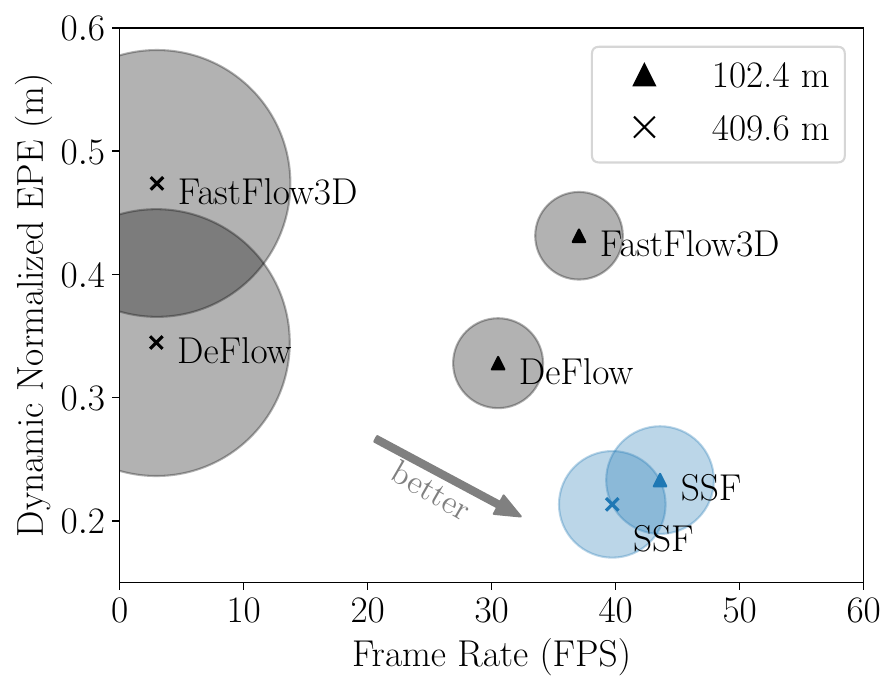}
\caption{
A plot of mean dynamic normalized EPE~\cite{khatri2024can} against the frame rate of inference for the validation set of Argoverse2 sensor dataset. $\blacktriangle$ and $\times$ represent the size of a square grid centered around ego-vehicle, which is used for feature extraction and training of the neural network. As the grid size is increased, the inference memory of dense \gls{bev} based methods increases, as indicated by the size of translucent circles around the markers. Our method, SSF demonstrates state-of-the-art performance while maintaining low memory and runtime.}
\label{fig:epe_vs_fps}
\vspace{-1.0em}
\end{figure}

In recent years, spatially sparse convolutions~\cite{graham2014spatially} have successfully enhanced the computational efficiency, memory usage, and latency of 3D object detection networks. 
Consequently, they have also been used to improve long-range performance of \gls{lidar} based 3D object detection networks. The key is to allocate features only for occupied cells, which makes the computation scale roughly with the number of \gls{lidar} points and their distribution, rather than range and resolution as a \gls{bev} representation does. 
In this work, we show how to incorporate sparse convolutions to scene flow estimation. For accurate scene flow, the ability to fuse features from time-sequential point clouds is crucial. 
However, sparse convolution-based methods face a challenge in dynamic scenes: the number of points and their spatial distribution vary per scan, resulting in inconsistent feature vector sizes across time steps.

To overcome this drawback, we propose a novel feature fusion scheme, that preserves the ordering of grid features, ensuring consistent feature fusion. Equipped with this ability of fusing sparse feature vectors, we follow the design principles laid down in existing exemplar methods, and propose \gls{ssf}. Our method demonstrates state-of-the-art results on the \gls{av2} scene flow benchmark.
Furthermore, we observe that the computational efficiency afforded by the sparse backbones allows us to increase the range of our scene flow prediction while maintaining accuracy and a low memory footprint, as shown in~\Cref{fig:epe_vs_fps}. 
To the best of our knowledge, we are the first to report results for scene flow beyond 50 meters range. 
As mentioned previously, since existing scene flow benchmarks only focus on points close to the ego-vehicle, their evaluation metrics do not focus adequately on the impact of object range on scene flow. To address this gap, we propose a new metric, \textit{Range-wise EPE}, that integrates range sensitivity into the evaluation of scene flow.
The strong performance of SSF on this metric further validates the effectiveness of our method for long-range scene flow.

To summarize, in this work we make the following contributions: 

\begin{itemize}
    \item We propose \gls{ssf}, a long-range method for scene flow incorporating sparse convolution based backbone.
    \item We propose to overcome the drawback of varying feature size per scan through a simple scheme to allow sparse fusion of time-sequential data.
    \item We propose a novel range-wise metric and show that our method, SSF, outperforms existing methods on the task of long-range scene flow, while maintaining a low computational budget.
    \item We achieve the state-of-the art scene flow estimation performance on \gls{av2} dataset.
\end{itemize}

\section{Related Work} \label{sec:related_work}

\subsection{Scene Flow in Autonomous Driving}
Scene flow was presented as a generalized 3D counterpart of optical flow in an early work by Vedula et. al.~\cite{vedula1999three}.  
To handle much larger point clouds in the real-world (100K points or more), FastFlow3D~\cite{jund2021scalable} adopts the pillar based encoder~\cite{lang2019pointpillars}, which rasterizes the environment into a \gls{bev} grid and uses the points inside a cell to extract grid features. This 2D feature grid is processed akin to an image using \gls{cnn} layers, by a U-Net autoencoder~\cite{ronneberger2015u} to extract multiscale features. The resulting grid features are assigned back to points, and a linear decoder composing of a \gls{mlp} outputs the scene flow. Owing to its ability in processing large-scale point clouds, the architecture proposed in~\cite{jund2021scalable} is widely adapted in subsequent works. DeFlow~\cite{zhang2024deflow} further extract finegrained point-wise features by replacing the linear decoder with a specialized \gls{gru} based decoder. 
As opposed to pillar based encoders, VoxelNet~\cite{zhou2018voxelnet} creates a sparse volumetric feature grid and processes it using 3D convolutions for the task of 3D object detection. 
However, the computational complexity of dense 3D convolutions scales cubically with both voxel resolution and grid size, whereas pillar-based encoders scale quadratically with these factors.
This rapid increase in computational demand prohibits scalability, especially as the range of point clouds increases.

Another line of work treats scene flow as a runtime optimization problem. Notably, NSFP~\cite{li2021neural} proposes to use a simple \gls{mlp} as an implicit regularizer to obtain smooth flow for large-scale point clouds. More recent optimization-based works~\cite{li2023fast,li2024fast} strive to decrease the runtime cost through correspondence free loss functions, or embedding point features through a kernel function of support points. 
A recently proposed work~\cite{vidanapathirana2024multi} uses clustering to impose isometry in flow within rigid bodies. 
Although their runtime still limits applicability in a real-time system, the priors provided by optimization based methods have been successfully used in autolabeling~\cite{najibi2022motion, baur2024liso} and self-supervised scene flow learning~\cite{vedder2023zeroflow,zhang2024seflow}.  

\subsection{Sparse Methods for 3D Perception}
To address the challenge of scalability with respect to voxel resolution and range, recent research has focused on leveraging the inherent sparsity of point cloud data. Sparse 3D convolutions introduced by Graham et al.~\cite{graham2014spatially}, offer a more efficient approach by processing only the non-empty regions of the voxel grid. \gls{ssc}~\cite{graham2017submanifold} builds on this concept by avoiding activation at non-empty spatial locations, maintaining the sparsity pattern and thereby further improving the computational efficiency. \gls{ssc} has been successfully applied to 3D semantic segmentation~\cite{graham20183d} and 4D spatio-temporal processing~\cite{choy20194d}. Similarly, the SECOND~\cite{yan2018second} framework draws inspiration from VoxelNet's architecture, but enhances it by replacing the dense 3D convolutions with \gls{ssc}. 

Recent works~\cite{fan2022fully,chen2023voxelnext} have further refined SECOND's pipeline to make it fully sparse and demonstrated good results on long-range 3D object detection.
Due to the potential of these methods, in this paper we adopt SSC as the backbone of our approach to enhance the range of scene flow without directly increasing the computational burden.

\subsection{Metrics for Scene Flow}
The \gls{epe} has been a widely used metric for scene flow~\cite{liu2019flownet3d, wu2020pointpwc, gu2019hplflownet}. It measures the average Euclidean distance between predicted and ground truth flow vectors, providing a straightforward quantification of error in physical units (e.g. meters). Chodosh et. al.~\cite{Chodosh_2024_WACV} observed that the \gls{epe} is heavily influenced by background points which are often static. To address this limitation, they proposed the three-way \gls{epe}, which divides the point cloud into three parts: \gls{bs}, \gls{fs} and \gls{fd}. By averaging the \gls{epe} for each of these parts, the metric assigns higher weightage to foreground and dynamic points. 

The bucket-normalized \gls{epe} proposed in~\cite{khatri2024can}  provides a more granularized evaluation of scene flow across different categories and motion profiles. The authors categorize points into various classes and speed buckets, with an interval of 0.4~\si{m/sec}. The points with speed between 0-0.4~\si{m/sec} are categorized as static and their average across classes gives the static \gls{epe}. For the remaining buckets the average \gls{epe} is divided by the average speed of that bucket, then averaged across all non-empty buckets for each class, to yield the dynamic normalized \gls{epe}.



\section{Problem Statement} \label{sec:prob_statement}

\begin{figure*}[ht]
\centering
\includegraphics[trim=0 50 0 140, clip, width=\linewidth]{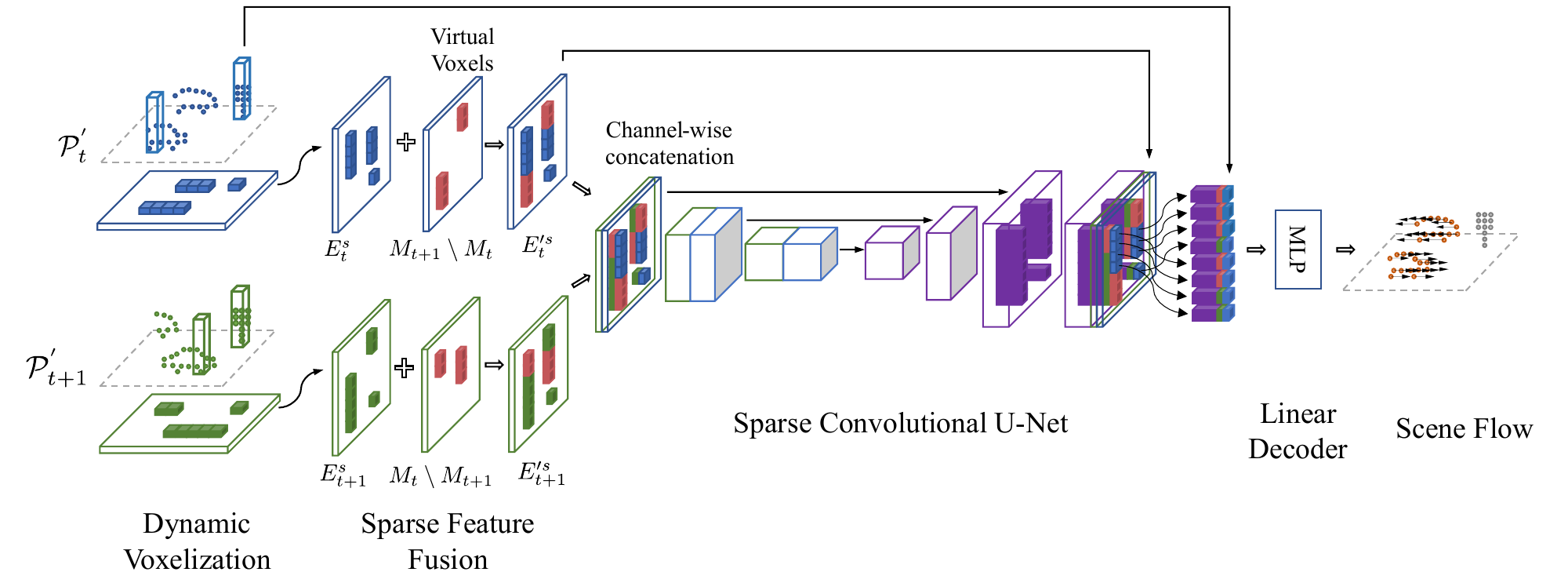}
\caption{
Schematic of our SSF model. The network takes as input point clouds at time t and t+1, shown in blue and green respectively. The first step involves voxelizing the space into pillars and computing sparse voxel feature encoding (VFE) features $E^s_t$ and $E^s_{t+1}$. Here, $M_t$ and $M_{t+1}$ denote the masks of voxels occupied by the point clouds at times t and t+1. To facilitate fusion by concatenation, the VFE feature maps are augmented with virtual voxels at locations defined by the set differences $M_{t+1} \setminus M_t$ and $M_t \setminus M_{t+1}$. The concatenated sparse feature maps are then processed using a sparse U-Net autoencoder. Finally, a linear decoder combines the encoder's output with sparse feature maps and point-wise offsets to generate per-point scene flow, represented by the arrows in the rightmost image.}
\label{fig:main_schematic}
\vspace{-1.0em}
\end{figure*}

Given two point clouds $\mathcal{P}_{t}$ and $\mathcal{P}_{t+1}$ at time $t$ and $t+1$ respectively, the problem of scene flow involves estimating a 3D vector $\mathcal{\Hat{\mathcal{F}}}_{t,t+1}$ which is applied to each point $\mathbf{p}_t=(x_t,y_t,z_t)$ in $\mathcal{P}_t$ to get a predicted point cloud $\Hat{\mathcal{P}}_{t+1}$. The goal is to minimize the \gls{epe} between predicted flow $\mathcal{\hat{F}}_{t,t+1}$ and ground truth flow $\mathcal{F}^*_{t,t+1}(p)$:
%
\begin{equation}
    \text{EPE}(\mathcal{P}_t) = \frac{1}{|\mathcal{P}_t|} \sum_{p \in \mathcal{P}_t} \left\| \mathcal{\hat{F}}_{t,t+1}(p) - \mathcal{F}^*_{t,t+1}(p) \right\|_2 \label{eq:epe}
\end{equation}
where $|\mathcal{P}_{t}|$ denotes the number of points in scan $\mathcal{P}_{t}$. In addition to time-sequential point clouds,
the inputs also consist of ground point masks $\mathcal{G}_t$ and $\mathcal{G}_{t+1}$ at time $t$ and $t+1$
The ground masks can be provided by \gls{hd} maps, or obtained using ground segmentation algorithms~\cite{steinke2023groundgrid}. The flow vector can be decomposed into two parts:
\begin{gather}
    \mathcal{\hat{F}}_{t,t+1} = \mathcal{F}_{ego} + \Delta \mathcal{\hat{F}}_{t,t+1} 
\end{gather}
where $\mathcal{F}_{ego}$ is produced by ego-motion $\mathbf{T_{t,t+1}}$ between time $t$ and $t+1$. $\mathbf{T_{t,t+1}}$ is assumed available and accurate, and is used as input. The residual flow due to motion of other agents, is estimated using a 
learned function $\mathcal{H}_\theta$ taking time-sequential non-ground point clouds as input:
\begin{gather}
    \Delta \mathcal{\hat{F}}_{t,t+1} = \mathcal{H}_\theta \left( \mathcal{P}^{'}_t, \mathcal{P}^{'}_{t+1} \right)
\end{gather}
where
\begin{gather}
    \mathcal{P}^{'}_t = \mathbf{T_{t,t+1}} (\mathcal{P}_t \setminus \mathcal{G}_t)\\ 
    \mathcal{P}^{'}_{t+1} = \mathcal{P}_{t+1} \setminus \mathcal{G}_{t+1} 
\end{gather}
Here $\mathcal{P}^{'}_t$ is the point cloud $\mathcal{P}_t$ with the ground points removed and transformed to $t+1$ using ego-motion.

\section{Sparse Scene Flow} \label{sec:method}
In the following section we outline our method, \gls{ssf}, which is divided into into following components: voxel encoding, sparse feature fusion, backbone and decoding. These components are described respectively in \crefrange{subsec:voxel_encoder}{subsec:backbone}. We follow the basic structure of FastFlow3D, but make critical changes to make the pipeline sparse and to ensure that feature ordering is preserved between the two scans in the sparse representation, as illustrated in~\Cref{fig:main_schematic}.







\subsection{Voxel Encoder} \label{subsec:voxel_encoder}
The first step involves rasterizing $\mathcal{P}^{'}_t$ and $\mathcal{P}^{'}_{t+1}$ within a square grid of size $R \times R$ centered on the ego-vehicle, and with a voxel resolution $(v_x, v_y, v_z)$, to get per-point voxel coordinates. For a pillar grid, $v_x=v_y$ and $v_z$ is chosen to be larger compared to $v_x$ and $v_y$. Notably, the points outside the grid are removed from processing, leading to a perception range of $R/2$. Next, all points within a particular grid cell are grouped together and augmented with additional features in form of the offset from voxel center $(\Delta o_x, \Delta o_y, \Delta o_z)$, and the cluster center $(c_x, c_y, c_z)$. This results in a 9-dimensional point feature vector, which is processed by an encoding layer. Here we use the dynamic \gls{vfe} layer from~\cite{fan2022fully} for its efficient implementation. Consisting of two layers of a \gls{mlp}, batch normalization and ReLU activation, the layer projects the point features to a higher dimensional space, which are then efficiently grouped into voxel features utilizing dynamic pooling. 

The dynamic \gls{vfe} encoder avoids the dense scattering operation employed by~\cite{lang2019pointpillars}, yielding a sparse feature map $E^s \in \mathbb{R}^{S\times C}$. In contrast the pillar based encoder used by other scene flow methods creates a dense 2D grid feature map $E^d \in \mathbb{R}^{D\times D\times C}$. Here $D=R/v_x$ corresponds to the size of the 2D grid, $S$ is the number of occupied voxels for a scan $\mathcal{P}$, and $C$ is the feature dimension. Crucially, $S << D\times D$, i.e., the number of actually occupied voxels is much smaller than the number of voxels in a dense $D\times D$ grid, which leads to a memory efficient sparser representation.








\subsection{Sparse Feature Fusion} \label{subsec:fusion}
Having obtained sparse feature maps $E^s_t \in \mathbb{R}^{S_0\times C}$ and $E^s_{t+1} \in \mathbb{R}^{S_1\times C}$ for scans $\mathcal{P}^{'}_t$ and $\mathcal{P}^{'}_{t+1}$, the next step is channel-wise fusion and feeding the result to a U-Net encoder, following~\cite{jund2021scalable}. 
However, a significant challenge is encountered: the sparse feature maps differ in size, making direct concatenation impossible. In contrast, in dense feature maps, concatenation is trivial because the feature maps have a consistent and regular grid structure. This uniformity allows for straightforward concatenation along the channel dimension. However, in our case, the sparse nature of the feature maps precludes such a simple solution. One might consider zero-padding the sparse feature maps to make them the same size, but this approach does not preserve the spatial ordering of voxels, leading to incorrect concatenation. 

To address this challenge, 
we combine the two scans before the voxelization step, and create a set of voxels $V$ occupied by at least one point from one of the point clouds. However, feature extraction must still be performed for each point cloud individually to efficiently encode motion information. 
To achieve this, we define indicator masks $M_t$ and $M_{t+1}$ of voxels belonging to scans $\mathcal{P}_{t}$ and $\mathcal{P}_{t+1}$ respectively. We fill the set of voxels occupied by one scan but not the other with virtual points before processing through the \gls{vfe} layer. After generating the \gls{vfe} features, the indicator masks are used to set the features of the virtual voxels to zero, negating the effects of the virtual points. This ensures that the resulting sparse maps $E^{'s}_t$ and $E^{'s}_{t+1}$ have the same size. 
As the point clouds were voxelized together, the ordering of voxels in $E^{'s}_t$ and $E^{'s}_{t+1}$ is also preserved, allowing for their fusion through concatenation. This is shown in \cref{fig:main_schematic}.
\subsection{Backbone and Head} \label{subsec:backbone}
As mentioned previously, existing works process the concatenated \gls{bev} feature maps using a UNet encoder-decoder, employing 2D convolutions. But in absence of dense feature maps, we replace the UNet from~\cite{jund2021scalable,graham20183d} with a sparse UNet from~\cite{fan2022fully,shi2019part}. 
Each layer in the encoder 
is composed of sparse 3D convolutions followed by two sub-layers of 3D \gls{ssc}. The encoder progressively decreases the spatial dimension and doubles the feature dimension, capturing more complex features at deeper layers. 
In the decoder, the architecture utilizes a series of lateral, merge and up-sampling blocks. The lateral block processes the features from corresponding encoder layer through 3D \gls{ssc}, and merges it with the deeper encoder layer features through concatenation. This is followed by another layer of 3D \gls{ssc} on merged tensor, channel reduction of the merged tensor and element-wise addition of the above two quantities. The final stage of each decoder layer involves sparse inverse convolutions to progressively reconstruct the high-resolution output from the downsampled features.   

Lastly, the points inside each grid are assigned the resulting grid feature through the unpillaring operation~\cite{jund2021scalable}. The indicator mask $M_t$ is used to select non-virtual points in scan $P^{'}_t$. Thereafter the point-wise decoder features are concatenated with the \gls{vfe} features as well as point-offset features and passed through a \gls{mlp} to obtain point-wise flow for scan $P^{'}_t$. 

\newcommand{\blue}[1]{$_{\color{TableBlue}\downarrow #1}$}
\newcommand{\white}[1]{$_{\color{TableWhite}\downarrow #1}$}
\begin{table*}[h]
\centering
\def\arraystretch{1.2}
\caption{Comparison to state-of-the-art on the \gls{av2} test set. The best and second best results are highlighted in bold and underlined respectively. The evaluation range is $35$~\si{m}. SSF (ours) outperforms other published methods in terms of mean three-way \gls{epe} and dynamic normalized \gls{epe}.}
\label{tab:test_comparison}
\begin{tabular}{cc|cccc|cc}
\toprule
  \multirow{2}{*}{Methods} & \multirow{2}{*}{Supervised} & \multicolumn{4}{c|}{Three-way EPE} & \multicolumn{2}{c}{Bucket-Normalized EPE} \\
                           &            & Mean ↓       & EPE FD ↓ & EPE FS ↓ & EPE BS ↓ & Dynamic Mean ↓ & Static Mean ↓ \\ \hline
Ego Motion Flow & ~ & 0.1813 & 0.5335 & \textbf{0.0103} & \textbf{0.0000} & 1.0000 & \textbf{0.0068} 
\\ \hline
SeFlow~\cite{zhang2024seflow} & ~ & 0.0486 & 0.1214 & 0.0184 & 0.0060 & 0.3085 & \underline{0.0142} 
\\
ICP Flow~\cite{lin2024icp} & ~ & 0.0650 & 0.1369 & 0.0332 & 0.0250 & 0.3309 & 0.0271 
\\
ZeroFlow~\cite{vedder2023zeroflow} XL 5x & ~ & 0.0494 & 0.1177 & \underline{0.0174} & 0.0131 & 0.4389 & 0.0143 
\\
NSFP~\cite{li2021neural} & ~ & 0.0606 & 0.1158 & 0.0316 & 0.0344 & 0.4219 & 0.0279 
\\
FastNSF~\cite{li2023fast} & ~ & 0.1118 & 0.1634 & 0.0814 & 0.0907 & 0.3826 & 0.0735 
\\ 
TrackFlow~\cite{khatri2024can} & \checkmark & 0.0473 & 0.1030 & 0.0365 & \underline{0.0024} & \underline{0.2689} & 0.0447 
\\
DeFlow~\cite{zhang2024deflow} & \checkmark & \underline{0.0343} & \underline{0.0732} & 0.0251 & 0.0046 & 0.2758 & 0.0218 
\\
FastFlow3D~\cite{jund2021scalable} & \checkmark & 0.0620 & 0.1564 & 0.0245 & 0.0049 & 0.5323 & 0.0182 
\\
SSF (Ours) & \checkmark & \textbf{0.0273} & \textbf{0.0572} & 0.0176 & 0.0072 & \textbf{0.1808} & 0.0154 
\\
\bottomrule
\end{tabular}
\end{table*}

\section{Range-wise \gls{epe}} \label{sec:range_epe}
In autonomous driving scenarios, the accuracy of scene flow estimation can significantly degrade with increasing distance from the sensor. This degradation is due to factors such as reduced point density, increased noise, and occlusions. Current metrics like the three-way EPE and bucket-normalized EPE do not explicitly incorporate range as a variable, potentially overlooking critical performance variations across different distances.


Inspired by~\cite{khatri2024can}, we divide our point cloud into bins according to point distance from the sensor. As reported in~\cite{Chodosh_2024_WACV}, the number of static points dominates the distribution compared to dynamic points, and we observed this to hold across all range bins. As the focus of the scene flow task is on the dynamic points, we further divide the points into two classes: static and dynamic. In contrast to~\cite{Chodosh_2024_WACV}, we classify points as dynamic using a threshold based on the normal walking speed of pedestrians, set at 1.4 m/sec.
Thereafter, for each range and class bin, we compute the \gls{epe} between the predicted flow and the ground truth. The range-wise \gls{epe} has two components: mean static and mean dynamic. These are obtained by averaging the \gls{epe} between all range bins for both static and dynamic classes respectively:
\begin{align}
    \text{Range-wise EPE}_{\text{static}} &= \frac{1}{k} \sum_{i=1}^k  \text{EPE}(B_i \cap \mathcal{S}) \label{eq:range_epe_static}
    \\
    \text{Range-wise EPE}_{\text{dynamic}} &= \frac{1}{k} \sum_{i=1}^k  \text{EPE}(B_i \cap \mathcal{D}) \label{eq:range_epe_dynamic}
\end{align}
where $B_i$ is a subset of points in range bin $i$, $k$ is the total number of bins, and $\mathcal{S}$ and $\mathcal{D}$ are the subsets of static and dynamic points respectively. EPE($\mathcal{P}$) is defined according to \cref{eq:epe}.

\section{Experiments}   \label{sec:expt}
\subsection{Experimental Setup}   \label{subsec:expt_setup}
\textbf{Dataset}: We perform our experiments on the  Argoverse2 (AV2)~\cite{wilson2argoverse} Sensor dataset.
The dataset consists of 1000 scenes of 15 seconds duration, divided into 700 scenes for training and 150 scenes each for the validation and testing. 

\textbf{Baselines and Metrics}: We first compare SSF to other published methods, including optimization based, as well as self-supervised learning based, on the \gls{av2} test set, obtained by uploading results to their evaluation server. 
The leaderboard uses two metrics for evaluation: bucket-normalized \gls{epe} \cite{khatri2024can} and three-way \gls{epe}~\cite{Chodosh_2024_WACV}, which computes the unweighted average of EPE on foreground dynamic (FD), foreground static (FS), and background static (BS). 
Next, we perform scalability experiments with respect to voxel resolution and perception range. We mainly compare against supervised dense \gls{bev} based scene flow methods FastFlow3D~\cite{jund2021scalable} and DeFlow~\cite{zhang2024deflow} on the \gls{av2} validation set. For long-range evaluation, we use our range-wise \gls{epe}. We additionally measure the memory usage and runtime, recorded in frames per second (FPS).

\textbf{Implementation Details}: Our \gls{ssf} implementation builds on DeFlow~\cite{zhang2024deflow}. 
For leaderboard comparison on the test set, evaluations are limited to a $70\times70$~\si{m} grid (or $35$~\si{m} range) around ego vehicle. Consequently, we initially train \gls{ssf} with grid size of $102.4\times102.4$~\si{m}, corresponding to a perception range of $51.2$~\si{m}. In our best-performing configuration, the pillar resolution $(v_x, v_y, v_z)$ is set to $(0.1, 0.1, 6)$~\si{m}. The learning rate is set to $8\times 10^{-3}$, with a batch size of $384$. We observe that increasing the learning rate significantly boosts the performance of FastFlow3d and DeFlow. Therefore, to ensure a fair comparison, we re-train both the models with higher learning rates during scalability experiments. All training runs are conducted for 25 epochs on A100 GPUs. The runtime experiments are performed on a desktop computer with an Intel i7-12700KF processor and a single RTX 3090 GPU.

\begin{table}[h]
\centering
\def\arraystretch{1.2}
\setlength{\tabcolsep}{3.0pt}
\caption{Analyzing the effect of voxel size on scene flow performance. The grid size is set to $102.4 \times 102.4$~\si{m}. The perception and evaluation range is 51.2~\si{m}. FPS denotes the frame rate for inference. As the voxel size is reduced, the memory goes up and frame rate reduces drastically for FastFlow3D and DeFlow. On the other hand, our method, SSF, maintains a constant memory and frame rate.}
\label{tab:voxel_size_abridged}
\begin{tabular}{cc|cc|c|c}
\toprule
  \multirow{3}{*}{Methods} & \multirow{3}{*}{Voxel size} & \multicolumn{2}{c|}{Bucket-Normalized EPE} & \multirow{3}{*}{Memory} & \multirow{3}{*}{FPS ↑} \\
                           &    &      Dynamic & Static  &   &  \\ 
                           &    (\si{m})  &      Mean ↓ & Mean ↓ &  (MB) ↓ &  \\ 
\hline
\multirow{2}{*}{FastFlow3D~\cite{jund2021scalable}} & 0.2   & 0.4314  & 0.0160 & 1736 & 38.26  \\

& 0.1    & 0.3389 & 0.0152 & 4616 & 12.73   
\\

\hline 

\multirow{2}{*}{DeFlow~\cite{zhang2024deflow}}     & 0.2  & 0.3279 & 0.0186 & 1732 & 30.52  
\\

& 0.1   & 0.2782 & 0.0185 & 4620 & 11.71   \\

\hline

\multirow{2}{*}{SSF (Ours)}        & 0.2   & 0.2330 & 0.0149 & 2610 & 43.58  
\\

& 0.1   & \textbf{0.1962} & \textbf{0.0137} & 2610 & 43.25  
\\

\bottomrule
\end{tabular}
\end{table}

\begin{table*}[h!]
\centering
\def\arraystretch{1.2}
\caption{Long-range scene flow evaluation using range-wise \gls{epe}. For all methods, the grid size is set to $409.6 \times 409.6$~\si{m}, corresponding to a perception range of $204.8$~\si{m}. The voxel size is set to $(0.2, 0.2, 6)$~\si{m}. SSF achieves the lowest mean dynamic error for all range bins.}
\label{tab:long_range}
\begin{tabular}{c|cccccc|cccccc}
\toprule
  \multirow{3}{*}{Methods} & \multicolumn{12}{c}{Range-wise \gls{epe}} \\
  & \multicolumn{6}{c|}{Dynamic ↓} & \multicolumn{6}{c}{Static ↓} \\
& 0-35 & 35-50 & 50-75 & 75-100 & 100+ & Mean & 0-35 & 35-50 & 50-75 & 75-100 & 100+ & Mean  \\ 
\hline
Ego Motion Flow & 0.6877  & 0.7365 & 0.7402 & 0.7008 & 0.5602  & 0.6851 & \textbf{0.0081} & \textbf{0.0085} & \textbf{0.0085} & \textbf{0.0091} & \textbf{0.0106}  & \textbf{0.0090} \\
  \hline
FastFlow3D~\cite{jund2021scalable}            & 0.1721  & 0.2630  & 0.3235 & 0.3777 & 0.3621  & 0.2997 & 0.0113 & 0.0123 & 0.0132 & 0.0147 & 0.0189  & 0.0141 \\
DeFlow~\cite{zhang2024deflow}                & \underline{0.1300}    & \underline{0.1984} & \underline{0.2444} & \underline{0.2937} & \underline{0.3356}  & \underline{0.2404} & 0.0097 & 0.0104 & 0.0112 & \underline{0.0131} & \underline{0.0179}  & \underline{0.0125} \\
SSF (Ours)                  & \textbf{0.1155}  & \textbf{0.1681} & \textbf{0.2045} & \textbf{0.2413} & \textbf{0.3054}  & \textbf{0.2069} & \underline{0.0084} & \underline{0.0091} & \underline{0.0111} & 0.0147 & 0.0197  & 0.0126 \\
\bottomrule
\end{tabular}
\end{table*}

\subsection{State-of-the-art Comparison} \label{subsec:quant_results}
We compare our method, \gls{ssf} to state-of-the-art approaches on the \gls{av2} scene flow leaderboard, as shown in \Cref{tab:test_comparison}. Our method gives the best performance on the metrics with dynamic components, with reduction of $21.8\%$ and $32.8\%$ respectively in \gls{epe} FD and mean dynamic \gls{epe} compared to next best method, DeFlow. For the static component, the \gls{epe} FS improves, but \gls{epe} BS increases slightly compared to DeFlow, indicating that \gls{ssf} is better in reasoning about the motion of foreground parts of the scene, but incurs some errors for background points. Overall, the three-way \gls{epe} also goes down by $20.4\%$ with a 1 centimetre error in static flow estimation.

The best performance on metrics with static components is observed for \gls{emc}. 
This is expected since the \gls{emc} only applies the motion vector corresponding to the ego-vehicle's movement, typically given by a localization system. 
As such, it represents a baseline with zero per-point dynamic flow prediction and works exceedingly well for stationary parts of the scene, as shown by low values for static components of both \gls{epe} metrics. 

\subsection{Scalability Experiments}
Computational efficiency is crucial when scaling a scene flow method to long-range, as finer resolution and extended perception range require larger networks, significantly increasing memory consumption. Two key design variables that impact the computational efficiency of the network are voxel resolution and grid size.

\textbf{Effect of voxel resolution}: 
\Cref{tab:voxel_size_abridged} reports the effect of reducing the voxel resolution. The bucket normalized \gls{epe} is reported together with the memory consumption and runtime (frame rate) for two different voxel sizes: $0.2$~\si{m} (default) and $0.1$~\si{m}, with all perception range fixed to $102.4 \times 102.4$~\si{m}. 

As the voxel size is reduced, the dynamic flow estimation accuracy of all methods increases.
However, for the \gls{bev}-based methods, FastFlow3D and DeFlow, this reduction leads to significantly higher memory consumption and a lower frame rate.
This is because for \gls{bev}-based methods, the number of grid cells increases quadratically with the inverse of the voxel size. Each grid cell, including empty ones, store features that undergo dense 2D convolution operations, resulting in a drastic rise in memory usage. On the other hand, \gls{ssf} benefits from our sparse feature fusion and maintains a consistent memory consumption and runtime, as it allocates features only for occupied voxels, which remain fairly constant regardless of the voxel size.


\begin{figure}[ht]
\centering
\includegraphics[trim=0 0 0 0, clip, width=0.99\linewidth]{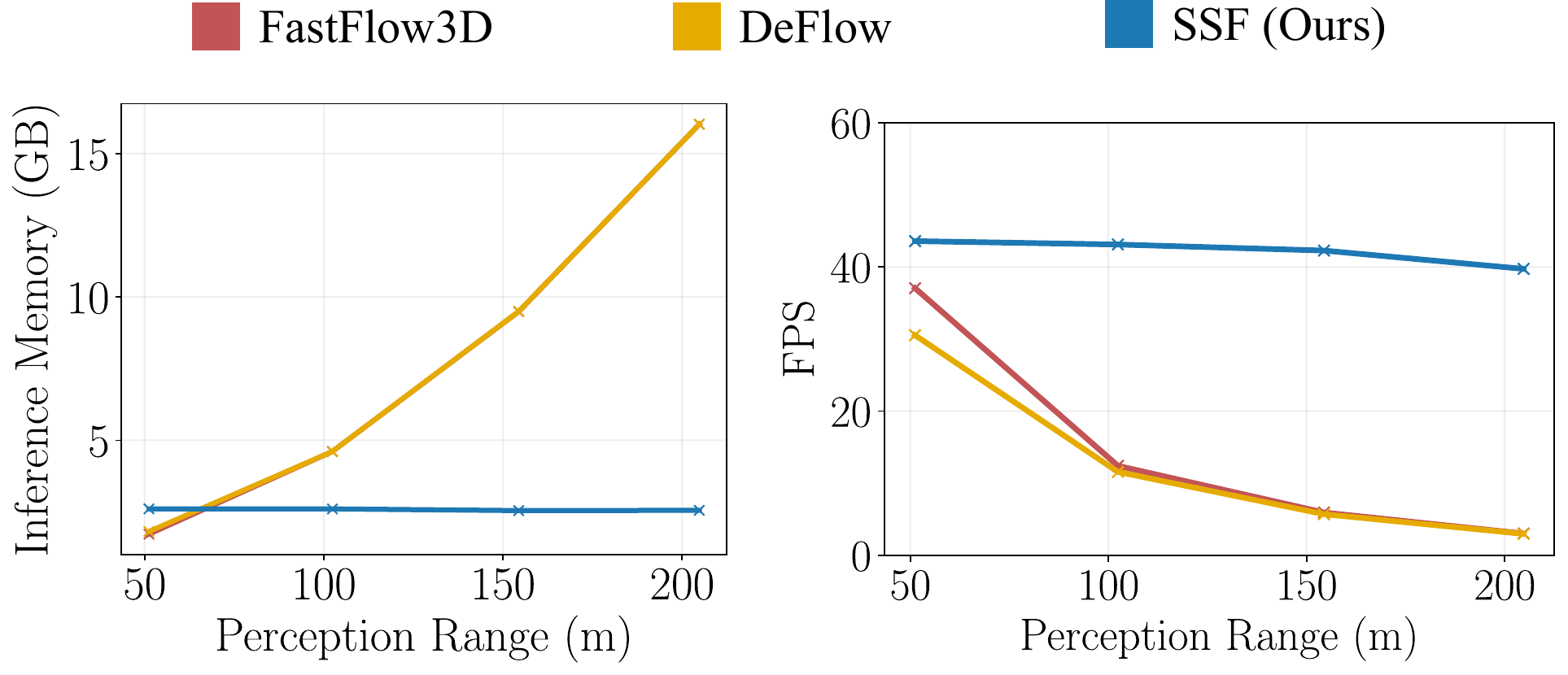}
\caption{Plots of inference memory and runtime against perception range.}
\label{fig:plots_vs_range}
\end{figure}

\textbf{Long-range Evaluation}: To make a fair evaluation on long-range, we fix the voxel size for all methods to a default of $0.2$~\si{m}. Thereafter we increase the grid size progressively and train the three methods for long-range scene flow. 
As the perception range increases, \gls{bev}-based methods consume more GPU memory and experience a decline in runtime (FPS), whereas \gls{ssf} maintains its performance with respect to both, as shown in \Cref{fig:plots_vs_range}. For a maximum grid size of $409.6\times409.6$~\si{m}, we report the performance using our range-wise \gls{epe} in \Cref{tab:long_range}. 
The range is categorized into bins with the following upper bounds: $35$~\si{m}, $50$~\si{m}, $75$~\si{m}, $100$~\si{m}, and infinity. The first bin size is chosen to align with the evaluation range used in the official leaderboard.

Here as well, we include the \gls{emc} as a baseline with low static error, but high dynamic error for each range-bin. \gls{ssf} maintains the best performance in terms of dynamic range-wise \gls{epe} across each range bin. Notably, \gls{ssf} reduces the mean dynamic range-wise \gls{epe} by $14\%$ compared to current state-of-the-art method DeFlow. Also, the \gls{epe} increases with range, highlighting a challenge which is overlooked by other existing metrics. As previously mentioned, this could be due to increased sparsity of points with range. Alongside better \gls{lidar} sensors, incorporating more time-sequential data can potentially improve the performance in the future.

\section{Conclusions and Future Work} 
\label{sec:conclusions}

In this paper, we propose SSF, a method for long-range scene flow for autonomous driving scenarios. SSF demonstrates performance comparable to the state-of-the-art of supervised scene flow benchmarks, while maintaining computational efficiency. We also propose a range-aware \gls{epe} metric to counter gaps in existing benchmarks in terms of long-range scene flow evaluation. Going forward, increased resolution and reduced costs will enable placement of multiple \gls{lidar}s on a vehicle while targeting different range profiles. We hope our contributions become more pertinent, and form a baseline for further developments in long-range perception. 

As future work, it would be interesting to extend our method to self-supervised learning and autolabeling of 3D objects. One could also incorporate scene flow backbones in existing 3D object detection approaches, providing priors in form of object hypotheses based on motion.






\section*{ACKNOWLEDGMENT}
\label{sec:acknowledgment}
This work was supported by the research grant PROSENSE (2020-02963) funded by VINNOVA. 
The computations were enabled by the supercomputing resource Berzelius provided by National Supercomputer Centre at Linköping University and the Knut and Alice Wallenberg Foundation, Sweden.

\bibliographystyle{IEEEtran}
\bibliography{mybib}

\begin{thebibliography}{10}
\providecommand{\url}[1]{#1}
\csname url@samestyle\endcsname
\providecommand{\newblock}{\relax}
\providecommand{\bibinfo}[2]{#2}
\providecommand{\BIBentrySTDinterwordspacing}{\spaceskip=0pt\relax}
\providecommand{\BIBentryALTinterwordstretchfactor}{4}
\providecommand{\BIBentryALTinterwordspacing}{\spaceskip=\fontdimen2\font plus
\BIBentryALTinterwordstretchfactor\fontdimen3\font minus \fontdimen4\font\relax}
\providecommand{\BIBforeignlanguage}[2]{{%
\expandafter\ifx\csname l@#1\endcsname\relax
\typeout{** WARNING: IEEEtran.bst: No hyphenation pattern has been}%
\typeout{** loaded for the language `#1'. Using the pattern for}%
\typeout{** the default language instead.}%
\else
\language=\csname l@#1\endcsname
\fi
#2}}
\providecommand{\BIBdecl}{\relax}
\BIBdecl

\bibitem{jund2021scalable}
P.~Jund, C.~Sweeney, N.~Abdo, Z.~Chen, and J.~Shlens, ``Scalable scene flow from point clouds in the real world,'' \emph{IEEE Robotics and Automation Letters}, vol.~7, no.~2, pp. 1589--1596, 2021.

\bibitem{zhang2024deflow}
Q.~Zhang, Y.~Yang, H.~Fang, R.~Geng, and P.~Jensfelt, ``{DeFlow}: Decoder of scene flow network in autonomous driving,'' in \emph{2024 IEEE International Conference on Robotics and Automation (ICRA)}, 2024, pp. 2105--2111.

\bibitem{vedder2023zeroflow}
K.~Vedder, N.~Peri, N.~Chodosh, I.~Khatri, E.~Eaton, D.~Jayaraman, Y.~Liu, D.~Ramanan, and J.~Hays, ``Zeroflow: Scalable scene flow via distillation,'' \emph{arXiv preprint arXiv:2305.10424}, 2023.

\bibitem{wilson2argoverse}
B.~Wilson, W.~Qi, T.~Agarwal, J.~Lambert, J.~Singh, S.~Khandelwal, B.~Pan, R.~Kumar, A.~Hartnett, J.~K. Pontes \emph{et~al.}, ``Argoverse 2: Next generation datasets for self-driving perception and forecasting,'' in \emph{Thirty-fifth Conference on Neural Information Processing Systems Datasets and Benchmarks Track (Round 2)}.

\bibitem{khatri2024can}
I.~Khatri, K.~Vedder, N.~Peri, D.~Ramanan, and J.~Hays, ``I can't believe it's not scene flow!'' \emph{arXiv preprint arXiv:2403.04739}, 2024.

\bibitem{graham2014spatially}
B.~Graham, ``Spatially-sparse convolutional neural networks,'' \emph{arXiv preprint arXiv:1409.6070}, 2014.

\bibitem{vedula1999three}
S.~Vedula, S.~Baker, P.~Rander, R.~Collins, and T.~Kanade, ``Three-dimensional scene flow,'' in \emph{Proceedings of the Seventh IEEE International Conference on Computer Vision}, vol.~2.\hskip 1em plus 0.5em minus 0.4em\relax IEEE, 1999, pp. 722--729.

\bibitem{lang2019pointpillars}
A.~H. Lang, S.~Vora, H.~Caesar, L.~Zhou, J.~Yang, and O.~Beijbom, ``Pointpillars: Fast encoders for object detection from point clouds,'' in \emph{Proceedings of the IEEE/CVF conference on computer vision and pattern recognition}, 2019, pp. 12\,697--12\,705.

\bibitem{ronneberger2015u}
O.~Ronneberger, P.~Fischer, and T.~Brox, ``U-net: Convolutional networks for biomedical image segmentation,'' in \emph{Medical image computing and computer-assisted intervention--MICCAI 2015: 18th international conference, Munich, Germany, October 5-9, 2015, proceedings, part III 18}.\hskip 1em plus 0.5em minus 0.4em\relax Springer, 2015, pp. 234--241.

\bibitem{zhou2018voxelnet}
Y.~Zhou and O.~Tuzel, ``Voxelnet: End-to-end learning for point cloud based 3d object detection,'' in \emph{Proceedings of the IEEE conference on computer vision and pattern recognition}, 2018, pp. 4490--4499.

\bibitem{li2021neural}
X.~Li, J.~Kaesemodel~Pontes, and S.~Lucey, ``Neural scene flow prior,'' \emph{Advances in Neural Information Processing Systems}, vol.~34, pp. 7838--7851, 2021.

\bibitem{li2023fast}
X.~Li, J.~Zheng, F.~Ferroni, J.~K. Pontes, and S.~Lucey, ``Fast neural scene flow,'' in \emph{Proceedings of the IEEE/CVF International Conference on Computer Vision}, 2023, pp. 9878--9890.

\bibitem{li2024fast}
X.~Li and S.~Lucey, ``Fast kernel scene flow,'' \emph{arXiv preprint arXiv:2403.05896}, 2024.

\bibitem{vidanapathirana2024multi}
K.~Vidanapathirana, S.-F. Chng, X.~Li, and S.~Lucey, ``Multi-body neural scene flow,'' in \emph{2024 International Conference on 3D Vision (3DV)}.\hskip 1em plus 0.5em minus 0.4em\relax IEEE, 2024, pp. 126--136.

\bibitem{najibi2022motion}
M.~Najibi, J.~Ji, Y.~Zhou, C.~R. Qi, X.~Yan, S.~Ettinger, and D.~Anguelov, ``Motion inspired unsupervised perception and prediction in autonomous driving,'' in \emph{European Conference on Computer Vision}.\hskip 1em plus 0.5em minus 0.4em\relax Springer, 2022, pp. 424--443.

\bibitem{baur2024liso}
S.~Baur, F.~Moosmann, and A.~Geiger, ``Liso: Lidar-only self-supervised 3d object detection,'' \emph{arXiv preprint arXiv:2403.07071}, 2024.

\bibitem{zhang2024seflow}
Q.~Zhang, Y.~Yang, P.~Li, O.~Andersson, and P.~Jensfelt, ``{SeFlow}: A self-supervised scene flow method in autonomous driving,'' in \emph{European Conference on Computer Vision (ECCV)}.\hskip 1em plus 0.5em minus 0.4em\relax Springer, 2024, p. 353–369.

\bibitem{graham2017submanifold}
B.~Graham and L.~Van~der Maaten, ``Submanifold sparse convolutional networks,'' \emph{arXiv preprint arXiv:1706.01307}, 2017.

\bibitem{graham20183d}
B.~Graham, M.~Engelcke, and L.~Van Der~Maaten, ``3d semantic segmentation with submanifold sparse convolutional networks,'' in \emph{Proceedings of the IEEE conference on computer vision and pattern recognition}, 2018, pp. 9224--9232.

\bibitem{choy20194d}
C.~Choy, J.~Gwak, and S.~Savarese, ``4d spatio-temporal convnets: Minkowski convolutional neural networks,'' in \emph{Proceedings of the IEEE/CVF conference on computer vision and pattern recognition}, 2019, pp. 3075--3084.

\bibitem{yan2018second}
Y.~Yan, Y.~Mao, and B.~Li, ``Second: Sparsely embedded convolutional detection,'' \emph{Sensors}, vol.~18, no.~10, p. 3337, 2018.

\bibitem{fan2022fully}
L.~Fan, F.~Wang, N.~Wang, and Z.-X. Zhang, ``Fully sparse 3d object detection,'' \emph{Advances in Neural Information Processing Systems}, vol.~35, pp. 351--363, 2022.

\bibitem{chen2023voxelnext}
Y.~Chen, J.~Liu, X.~Zhang, X.~Qi, and J.~Jia, ``Voxelnext: Fully sparse voxelnet for 3d object detection and tracking,'' in \emph{Proceedings of the IEEE/CVF Conference on Computer Vision and Pattern Recognition}, 2023, pp. 21\,674--21\,683.

\bibitem{liu2019flownet3d}
X.~Liu, C.~R. Qi, and L.~J. Guibas, ``Flownet3d: Learning scene flow in 3d point clouds,'' in \emph{Proceedings of the IEEE/CVF conference on computer vision and pattern recognition}, 2019, pp. 529--537.

\bibitem{wu2020pointpwc}
W.~Wu, Z.~Y. Wang, Z.~Li, W.~Liu, and L.~Fuxin, ``Pointpwc-net: Cost volume on point clouds for (self-) supervised scene flow estimation,'' in \emph{Computer Vision--ECCV 2020: 16th European Conference, Glasgow, UK, August 23--28, 2020, Proceedings, Part V 16}.\hskip 1em plus 0.5em minus 0.4em\relax Springer, 2020, pp. 88--107.

\bibitem{gu2019hplflownet}
X.~Gu, Y.~Wang, C.~Wu, Y.~J. Lee, and P.~Wang, ``Hplflownet: Hierarchical permutohedral lattice flownet for scene flow estimation on large-scale point clouds,'' in \emph{Proceedings of the IEEE/CVF conference on computer vision and pattern recognition}, 2019, pp. 3254--3263.

\bibitem{Chodosh_2024_WACV}
N.~Chodosh, D.~Ramanan, and S.~Lucey, ``Re-evaluating lidar scene flow,'' in \emph{Proceedings of the IEEE/CVF Winter Conference on Applications of Computer Vision (WACV)}, January 2024, pp. 6005--6015.

\bibitem{steinke2023groundgrid}
N.~Steinke, D.~Goehring, and R.~Rojas, ``Groundgrid: Lidar point cloud ground segmentation and terrain estimation,'' \emph{IEEE Robotics and Automation Letters}, vol.~9, no.~1, pp. 420--426, 2023.

\bibitem{shi2019part}
S.~Shi, Z.~Wang, X.~Wang, and H.~Li, ``Part-aˆ 2 net: 3d part-aware and aggregation neural network for object detection from point cloud,'' \emph{arXiv preprint arXiv:1907.03670}, vol.~2, no.~3, 2019.

\bibitem{lin2024icp}
Y.~Lin and H.~Caesar, ``Icp-flow: Lidar scene flow estimation with icp,'' in \emph{Proceedings of the IEEE/CVF Conference on Computer Vision and Pattern Recognition}, 2024, pp. 15\,501--15\,511.

\end{thebibliography}

\end{document}